\def\year{2021}\relax
\documentclass[letterpaper]{article} % DO NOT CHANGE THIS
\usepackage{aaai21}  % DO NOT CHANGE THIS
\usepackage{times}  % DO NOT CHANGE THIS
\usepackage{helvet} % DO NOT CHANGE THIS
\usepackage{courier}  % DO NOT CHANGE THIS
\usepackage[hyphens]{url}  % DO NOT CHANGE THIS
\usepackage{graphicx} % DO NOT CHANGE THIS
\usepackage{natbib}  % DO NOT CHANGE THIS AND DO NOT ADD ANY OPTIONS TO IT
\usepackage{caption} % DO NOT CHANGE THIS AND DO NOT ADD ANY OPTIONS TO IT
\usepackage{diagbox}
\usepackage{amsmath}
\usepackage{amssymb}
\usepackage{amsthm}
\usepackage{mathdots}
\usepackage{caption}
\usepackage{subcaption}
\usepackage{color}
\usepackage{mathtools}
\usepackage{acro}
\usepackage{booktabs}
\usepackage{sistyle}
\usepackage{fontawesome}
\usepackage[export]{adjustbox}
\usepackage[switch]{lineno}
\SIthousandsep{,}

\newcommand{\ie}{\emph{i.e.},}
\newcommand{\eg}{\emph{e.g.},}

\DeclareMathOperator*{\E}{\mathbb{E}}
    
\DeclareAcronym{PXR}{
short=PXR,
long=pelvic X-ray
}
\DeclareAcronym{PACS}{
short=PACS,
long=picture archiving and communication system
}
\DeclareAcronym{FROC}{
short=FROC,
long=free‐response operating characteristic 
}
\DeclareAcronym{AUROC}{
short=AUROC,
long=area under receiver operating characteristic curve
}
\DeclareAcronym{ROC}{
short=ROC,
long=receiver operating characteristic
}
\DeclareAcronym{CAD}{
short=CAD,
long=computer aided diagnosis
}
\DeclareAcronym{CAM}{
short=CAM,
long=class activation map
}
\DeclareAcronym{GAP}{
short=GAP,
long=global average pooling
}
\DeclareAcronym{LSE}{
short=LSE,
long=log-sum-exp
}
\DeclareAcronym{CXR}{
short=CXR,
long=chest X-ray
}
\DeclareAcronym{CT}{
short=CT,
long=computed tomography
}
\DeclareAcronym{FPN}{
short=FPN,
long=feature pyramid network
}
\DeclareAcronym{FCN}{
short=FCN,
long=fully convolutional network
}
\DeclareAcronym{MSE}{
short=MSE,
long=mean squared error
}
\DeclareAcronym{KLD}{
short=KLD,
long=Kullback–Leibler divergence
}
\DeclareAcronym{BCE}{
short=BCE,
long=binary cross entropy
}

\makeatletter
\newcommand{\printfnsymbol}[1]{%
  \textsuperscript{\@fnsymbol{#1}}%
}
\makeatother

\title{A New Window Loss Function for Bone Fracture Detection and Localization \\ in X-ray Images with Point-based Annotation}

 \author {
     \thanks{X. Zhang and Y. Wang—Equal contribution.}
     \thanks{This work was done when X. Zhang was intern at PAII Inc.}
     Xinyu Zhang\textsuperscript{\rm 1},
     \printfnsymbol{1}
     Yirui Wang\textsuperscript{\rm 1},
     Chi-Tung Cheng\textsuperscript{\rm 2},
     Le Lu\textsuperscript{\rm 1 },
     Adam P. Harrison\textsuperscript{\rm 1 },
     Jing Xiao\textsuperscript{\rm 3 }, \\
     Chien-Hung Liao\textsuperscript{\rm 2},
     Shun Miao\textsuperscript{\rm 1 } \\
 }
\affiliations {
     \textsuperscript{\rm 1} PAII Inc., Bethesda, Maryland, USA  \\ 
     \textsuperscript{\rm 2} Chang Gung Memorial Hospital, Linkou, Taiwan, ROC \\
    \textsuperscript{\rm 3} Ping An Technology, Shenzhen, China 
 }

\begin{document}

\maketitle
%\linenumbers % 

\begin{abstract}
Object detection methods are widely adopted for computer-aided diagnosis using medical images. Anomalous findings are usually treated as objects that are described by bounding boxes. Yet, many pathological findings, \eg{} bone fractures, cannot be clearly defined by bounding boxes, owing to considerable instance, shape and boundary ambiguities. This makes bounding box annotations, and their associated losses, highly ill-suited. In this work, we propose a new bone fracture detection method for X-ray images, based on a labor effective and flexible annotation scheme suitable for abnormal findings with no clear object-level spatial extents or boundaries. Our method employs a simple, intuitive, and informative point-based annotation protocol to mark localized pathology information. To address the uncertainty in the fracture scales annotated via point(s), we convert the annotations into pixel-wise supervision that uses lower and upper bounds with positive, negative, and uncertain regions. A novel \textit{Window Loss} is subsequently proposed to only penalize the predictions outside of the uncertain regions. Our method has been extensively evaluated on \num{4410} pelvic X-ray images of unique patients. Experiments demonstrate that our method outperforms previous state-of-the-art image classification and object detection baselines by healthy margins, with an AUROC of $0.983$ and FROC score of $89.6\%$.

\end{abstract}

\section{Introduction}
\label{sec.intro}

Due to its efficiency, accessibility, and low cost, X-ray imaging is one of the most commonly performed diagnostic examinations in clinical practice. Conventional computer vision techniques have been extensively researched in the past few decades to develop \ac{CAD} solutions, aiming to increase image reading efficiency and reduce misdiagnoses risks. Current state-of-the-art \ac{CAD} systems widely adopt generic principles of image classification and object detection to identify and localize the anomalies. However, to provide both robust classification and localization in applications where pathologies have ambiguously defined extents, such as for \acp{PXR}, alternative solutions are needed. This is the topic of our work.

Spurred by the public release of large-scale datasets, \ac{CXR} \ac{CAD} has received a great deal of recent attention, with many prominent methods treating the problem as a multi-label image classification problem~\cite{rajpurkar2017chexnet,wang2017chestx8,baltruschat2019comparison}. These works typically report \ac{AUROC} scores, \eg{} $0.817$~\cite{baltruschat2019comparison} on the NIH-ChestXray14 dataset~\cite{wang2017chestx8} and $0.930$~\cite{pham2020interpreting} on the CheXPert dataset~\cite{irvin2019chexpert}. Yet, in clinical diagnostic workflows, localization of the detected anomalies is also needed both to interpret the results and for verifiable decision making. While \acp{CAM} can provide a degree of localization under classification setups, they have limited localization power and tend to miss abnormal sites when multiple pathological findings occur concurrently in an image. 

\begin{figure}[t]
    \centering
    \noindent\includegraphics[width=0.48\linewidth]{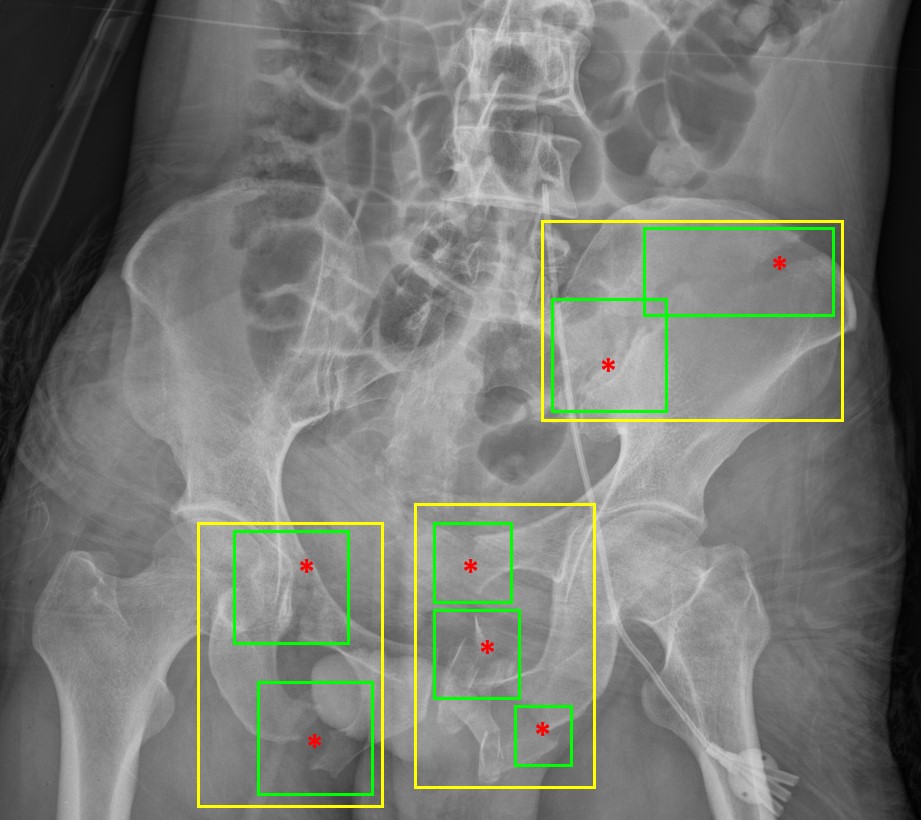} \includegraphics[width=0.48\linewidth]{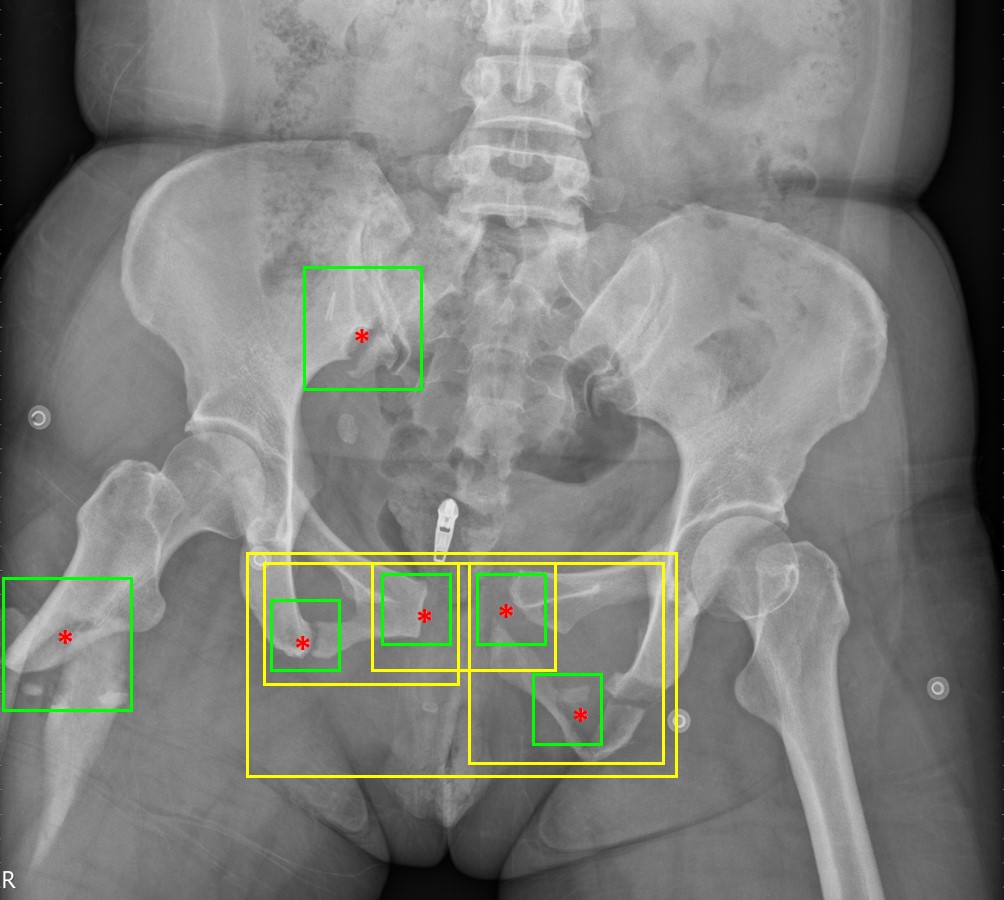} 
    \caption{Fractures in \acsp{PXR} where the numbers of instances and their spatial extents are extremely ambiguous. The annotations of fractures can be represented by small/tight bounding boxes on fracture lines (\textbf{green boxes}), or relatively big bounding boxes containing the entire fractured region (\textbf{yellow boxes}). To skirt this inherent subjectivity, we adopt and learn from point-based annotations (\textbf{red dots}). }
    \label{fig:intro}
    % \vspace{}
\end{figure}

To improve localization performance, several methods use additional box-level annotations to facilitate the training of image classifiers~\cite{li2018thoracic,liu2019align,huang2020rectifying}. For other \ac{CAD} applications where localization performance is more central, such as lung nodule and breast mass/calcification detection, explicit object detection frameworks are widely adopted~\cite{yan_2018_deeplesion,yan2019mulan,jiang2020elixirnet}. However, compared to image classifiers, object detection methods focus predominantly on localization and may struggle to produce robust image-level diagnoses, \eg{} by triggering undesirable false positives. 

Another critical downside to object detectors is their reliance on bounding boxes. Besides being prohibitively laborious, bounding boxes (or any explicit defining boundary) are not always suitable for all types of medical anomalies, particularly when pathological findings cannot be easily delineated as discrete instances with definite boundaries. Fractures in \acp{PXR} serve as excellent exemplar, as they usually involve multiple anatomical sites with complex morphology and curvilinear fragments that make defining the border, extent, and number of the bounding boxes extraordinarily difficult and ambiguous (as shown in Fig. \ref{fig:intro}). Therefore, an anomaly diagnosis and localization method with 1) a cost-effective and flexible annotation scheme and 2) simultaneously high classification and localization performance is critical and highly desirable.

In this work, we present a trauma \ac{PXR} bone fracture diagnosis and localization system with a cost-effective and flexible annotation scheme that is suitable for detecting abnormal findings with no clear object boundaries. Interpreting trauma \acp{PXR} in an emergency room environment requires timely image-level diagnoses and fracture localizations. The image-level diagnosis determines the downstream actions in the clinical workflow, \eg{} further examination, hospitalization, or hospital discharge. Robust fracture localization can help physicians avoid missed diagnoses that must be treated in a timely manner, otherwise serious complications can occur. Due to the geometrical complexities of bone fractures and the perspective projection distortions of \acp{PXR}, diagnosis and localization of fractures in \ac{PXR} is a challenging task where even experienced physicians can make considerable diagnostic errors~\cite{chellam2016missed}.

To cope with the inherent geometrical complexity of pelvic fractures, our model use annotated points as supervisory signals. This point-based annotation is labor efficient and allows ambiguous fracture conditions to be adequately represented as one point or multiple points (on the fracture sites). To account for the uncertainties in the scale and shape of the annotated fracture sites, we propose to convert the points to pixel-wise supervision signals, which consist of lower and upper bounds of the expected network prediction with an allowed region between the bounds. A novel \textit{Window Loss} is proposed that only penalizes predictions outside of the bounds, which encourages the model to learn from the reliable information without being distracted by the inherent uncertainties in the annotations. We evaluated our proposed method on \num{4410} \acp{PXR} collected from a primary trauma center representing real clinical situations. Our method reports an AUROC of $0.983$ and FROC score of $89.6\%$, significantly outperforming previous state-of-the-art image classification and object detection methods by at roughly $1\%$ or more on AUROC and $1.5\%$ on FROC score, respectively.

\section{Related Work}
\label{sec.related}

\subsection{Image Classification Methods}

The availability of large-scale public \ac{CXR} datasets with image-level labels has catalyzed a large body of work on \ac{CXR} classification. Typically, these methods train deep neural networks that output per-image classification scores and calculate losses against the image-level labels. \textit{CheXNet}~\cite{rajpurkar2017chexnet} trains a DenseNet-121~\cite{huang2017densely} with \ac{GAP} performed on the penultimate feature maps before being inputted into a fully connected layer, and \citet{wang2017chestx8} replaced the \ac{GAP} with \ac{LSE}. \citet{yao2018weakly, wang2019weakly,li2018thoracic,liu2019align} take a different approach by regarding the convolutional feature maps as local appearance descriptors of spatially sliced blocks. The features maps are then processed with a 1x1 convolution to produce class-wise probability maps and globally pooled to produce the image-level classification probabilities. Choices of pooling methods include \ac{LSE} pooling~\cite{yao2018weakly, wang2019weakly} or the product of the probabilities~\cite{li2018thoracic,liu2019align}, where the latter assumes that probability values are independent and identically distributed.

Methods with global feature pooling~\cite{yan2018weakly} typically rely on spatial attention mechanisms like \acp{CAM} or Grad-CAM to localize the underlying anomalies. In contrast, methods with class-wise probability maps~\cite{yao2018weakly, wang2019weakly} can directly produce localizations without any add-on components. Either way, deep models trained with only image-level labels may not always produce plausible localizations with high reliability. Other approaches~\cite{li2018thoracic,liu2019align} have addressed this issue by developing strategies to exploit the small number/percentage of images with bounding box annotations. 

Similar to prior work~\cite{chen2020anatomy}, our method uses a 1x1 convolution on the last feature map to produce a fracture class probability map. Different from the above classification based methods, our method is trained using point-based labels. In this way, we share a similar motivation with \citet{li2018thoracic,liu2019align} to leverage additional low cost localization annotations. However, we apply a fundamentally different conceptual framework, using  1) point-based annotations to account for the intrinsic ambiguity on visually defining/depicting bone fractures in X-rays as spatially-bounded objects and 2) designing a new Window Loss to naturally handle and learn from these annotations.

\begin{figure*}[!tb]
    \centering
    \includegraphics[width=\linewidth]{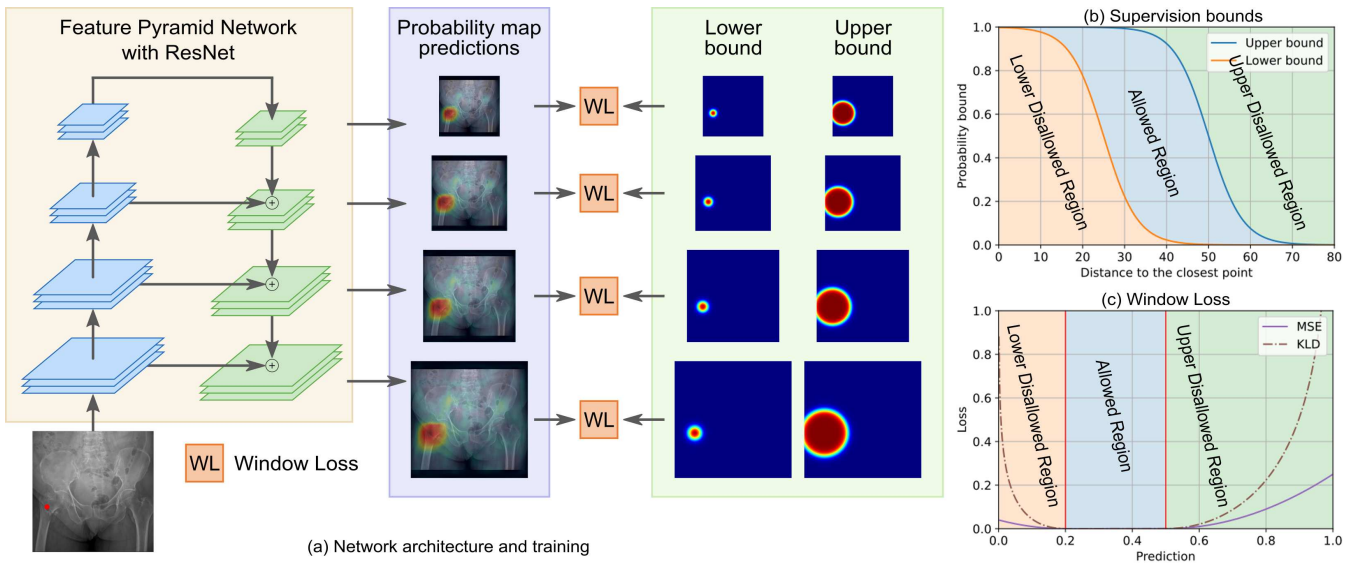}
    % \setlength{\tabcolsep}{0pt}
    % \renewcommand{\arraystretch}{0}
    % \begin{tabular}[t]{cc}
    % \begin{subfigure}{0.68\textwidth}
    %     \centering
    %     \smallskip
    %     \includegraphics[width=\linewidth, left]{network.png}
    %     \caption{Network architecture and training.} %{Light Unit}
    %     \label{fig:system}
    % \end{subfigure}
    % &
    % \begin{tabular}{c}% if you add [t], than sub images are pushed down
    % \smallskip
    %     \begin{subfigure}[t]{0.32\textwidth}
    %         \centering
    %         \includegraphics[width=0.95\textwidth, right]{window_loss_gt.eps}
    %         \caption{Supervision bounds.}
    %         \label{fig:supervision}
    %     \end{subfigure}\\
    %     \begin{subfigure}[t]{0.32\textwidth}
    %         \centering
    %         \includegraphics[width=0.95\textwidth, right]{window_loss.eps}
    %         \caption{Window Loss}
    %         \label{fig:wl_profile}
    %     \end{subfigure}
    % \end{tabular}\\
    % \end{tabular}
    \caption{\textbf{Framework of the proposed method}. (a) Network architecture and training mechanism of the proposed method. (b) Lower and upper bounds of the generated pixel-level supervision. (c) Profiles of the proposed Window Loss using MSE or KLD as the divergence measure. }
    \label{fig:system}
\end{figure*}

\subsection{Object Detection Methods}
% Object detection aims to predict a bounding box location and a category label for each instance of interest found in an image. 
% The current mainstream detectors are all designed in a fully convolutional setting, no matter in an anchor-based manner \cite{ren2015faster,liu2016ssd,redmon2016you}, or being anchor-free \cite{tian2019fcos, law2018cornernet}.
Substantial progress has been made in recent years in the development of deep learning based object detection methods. These include anchor-based solutions~\cite{ren2015faster,lin2017focal} and recent anchor-free solutions~\cite{tian2019fcos,law2018cornernet}, which repeatedly break the accuracy records on benchmarks like MS-COCO \cite{lin2014microsoft}. 
The success and popularity of object detection in natural imagery has motivated research on adopting them for a wide spectrum of medical image anomaly localization problems, such as distal radius fractures in hand X-rays~\cite{yahalomi2019detection}, pneumonia in \ac{CXR}~\cite{sirazitdinov2019deep}, masses in mammography~\cite{al2018simultaneous} and general lesions in \ac{CT}~\cite{yan_2018_deeplesion,yan2019mulan}.
Many domain specific enhancements have been proposed, including incorporating symmetry cues for breast mammography~\cite{liu2019unilateral}; exploiting the anatomical prior of vertebraes via self-calibration~\cite{zhao2019automatic}; and addressing missing annotations, hard negatives, and/or heterogeneous datasets for universal lesion detection~\cite{yan2020universal,cai2020lesion}. 
Like object detectors, our work explicitly learns from localization annotations, but the inherent visual ambiguity of pelvic fractures %with no clear spatial definition or even instance count
makes our problem fundamentally different. To address this, we propose a point-based annotation scheme and Window Loss for such challenging ambiguities.

\section{Method}
\label{sec.method}

There are two critical metrics to measure a \ac{CAD} solution's effectiveness in clinical settings: {\bf generality} (handling of unseen patients) and {\bf precision} (generating good accuracy robustly for the given task). Naturally, incorporating as many patients as possible for training is desirable, but patient data often cannot be fully annotated at scale, precluding approaches seen in natural images~\cite{Lin2014coco}. In medical imaging, the labor costs are always paramount. Even more intractably, for many applications, \eg{} \ac{PXR} fracture detection, there are inherent perceptual uncertainties and challenges in annotating precise bounding boxes of pathological findings. Alternative strategies are needed. A promising inspiration can be found in extreme points for object annotation~\cite{Papadopoulos2017Extreme}, which are found to be a quicker and easier visual perception protocol by human annotators. %We train a \ac{FCN} that takes as a given \ac{PXR} as input and outputs a dense probability map representing the confidence of the existence of fracture sites. 
Similarly, in this work, we use only point-based annotations to localize bone fractures in \acp{PXR} (\ie{} {\it where to look?}) so we can execute the annotation process at scale to cover as many as thousands of patients to achieve {\bf generality}. To achieve {\bf precision}, we propose a Window Loss that can robustly cope with the aforementioned ambiguities in defining the extents and scales of fracture sites (\ie{} {\it how to look?}). %We will describe the network architecture, point-based annotation and the Window Loss in the following sections, respectively. 
Fig. \ref{fig:system} illustrates an overview of our framework.

\subsection{Point-based Annotation}
\label{sec.method.annotation}

The complex morphology of pelvic fractures challenges the collection of reliable and consistent localization annotations. Due to the difficulties in defining the border, extent, and number of bounding boxes, the commonly adopted bounding box annotation suffers from high inter-annotator variations, which can cause conflicting supervisory signals. For instance, the pelvic fractures shown in Fig. \ref{fig:intro} can be annotated using multiple small boxes or one larger box. We propose to use point-based annotations to provide the localization information. The pin-pointing annotation protocol is flexible and naturally fits our complex scenario: we ask the annotators/physicians to place points on visible fracture sites. For complex scenarios where the instance of fractures cannot be clearly defined, the annotators decide to place one or multiple points at their own discretion. Two challenges arise when using point-based annotation to supervise network training: 1) unknown scales of fracture sites, and 2) inter-rater variability in point placements. We address this with our proposed Window Loss.

\subsection{Window Loss}
\label{sec.method.loss}

Despite any uncertainties in scale and inter-annotator variability, point-based annotations provide the following information: 1) regions in close proximity to the annotated points are likely abnormal; 2) regions far away from all points are highly likely to be normal; 3) for in-between regions it is difficult, if not possible, to specify any particular confidence value. We encode this by calculating an allowed confidence range at every point in the image. Any prediction that falls into the allowed range is deemed acceptable, regardless of where it falls. This aligns with the inherent ambiguities of fracture localization, where many image regions do not permit the specification of any particular value. On the other hand, if any prediction violates the allowed confidence range, it is penalized. 

To calculate an allowed confidence range, we create two heatmaps to represent the upper confidence and lower confidence bounds at every pixel location. Both heatmaps are set to $1$ at any point annotation and their values decline toward $0$ at points further away. However, the lower confidence bound reduces rapidly as the distance to the point increases, whereas the upper bound reduces more gradually. We construct both heatmaps using the sigmoid of the distance to the annotation points:

\begin{align}
    \ell(i, j) &= \max_k \sigma(\frac{r_{\ell}-D^k_{ij}}{\tau}), \\
    u(i, j) &= \max_k \sigma(\frac{r_{u}-D^k_{ij}}{\tau}),
\end{align}
where $\sigma$ is the sigmoid function, $D^k_{ij}$ denotes the distance from pixel $(i,j)$ to the $k$-th annotation point, and $\tau$ controls the softness of the confidence decline. Here the rate of the confidence bound decline is controlled by $r_{u}$ and $r_{\ell}$ for the upper and lower bounds, respectively. Importantly, $r_{u}>r_{\ell}$. Example profiles of the two confidence bounds are shown in Fig. \ref{fig:system}b. Regions above the upper bound or below the lower bound are disallowed regions, whereas regions between the two bounds are allowed ranges. 

The \textit{Window Loss} aims to discourage confidence predictions in any of the disallowed regions, while allowing any prediction within the allowed range. Dropping the pixel indexing for simplicity, we formulate the Window Loss as

\begin{equation}
    WL(p, \ell, u) = \begin{cases}
       D(p, \ell) & \text{if $p \leq \ell$},  \\
       0 & \text{if $\ell < p \leq u$}\\
       D(p, u) & \text{if $p > u$}  \end{cases},
\end{equation}
where $D(\cdot, \cdot)$ is a chosen divergence measure. Formulated this way, the Window Loss encouraged predictions violating the upper bound (resp., lower) to be lower (resp., higher), while not penalizing any predictions falling within the allowed regions. In term of divergence measures, we evaluate two options: \ac{MSE} and \ac{KLD}, written as

\begin{align}
    D_{MSE}(x, y) &= \| x - y \|^2, \\
    D_{KLD}(x, y) &= y\log(\frac{y}{x}) + (1-y)\log(\frac{1-y}{1-x}).
\end{align}
Examples of the Window Loss using \ac{MSE} and \ac{KLD} are shown in Fig.~\ref{fig:system}c. The Window Loss can garner significant improvements in performance where ambiguities are unavoidable. Its simple formulation allows it to be easily plugged into any existing deep learning framework.

\begin{figure*}[t]
    \centering
    \noindent\includegraphics[width=\linewidth]{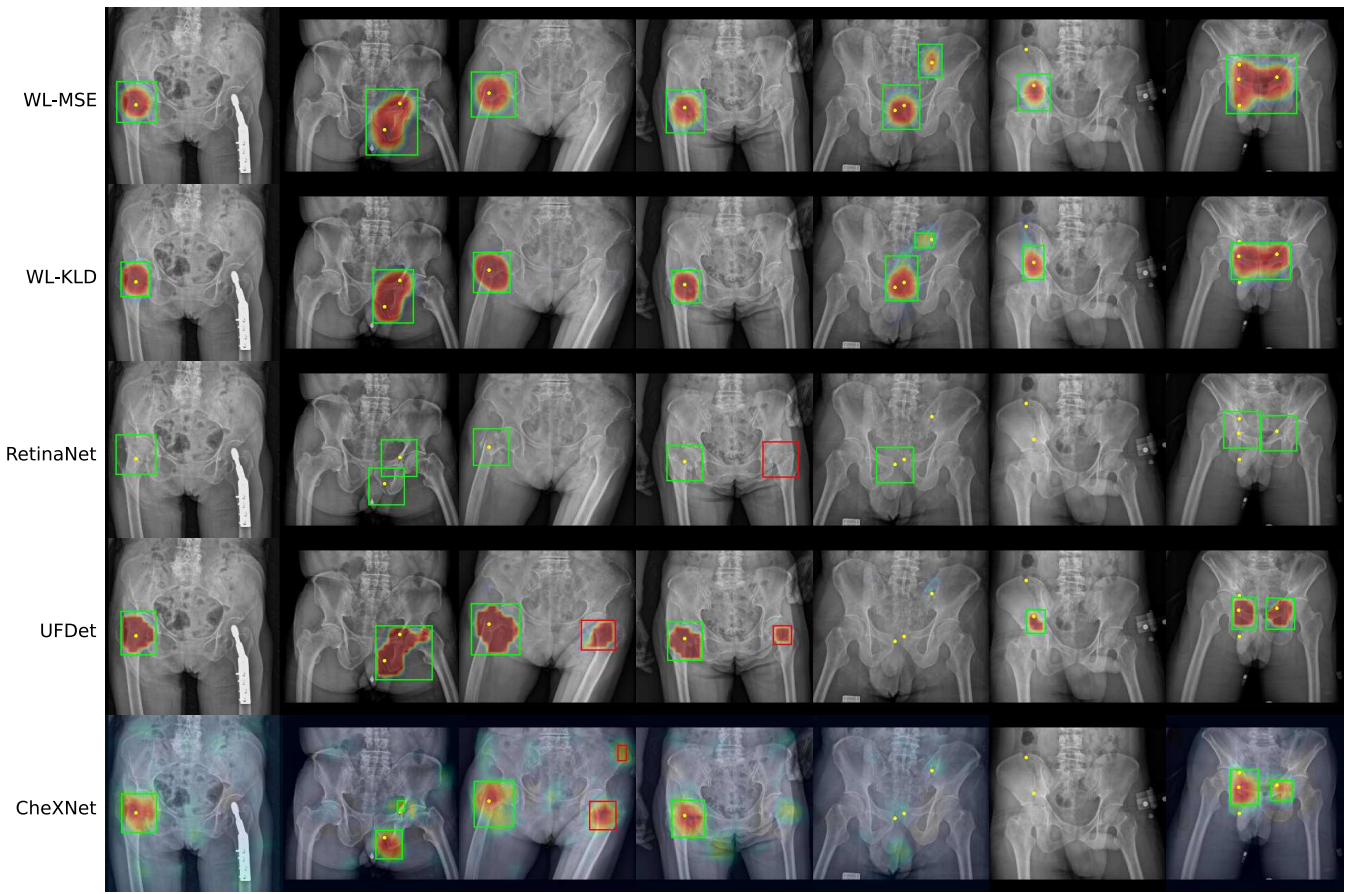} 
    \caption{\textbf{Visualization of fracture localization results}. The proposed methods, WL-MSE and WL-KLD, achieve higher localization performances than the previous state-of-the-art methods. The localization performance of WL-MSE is qualitatively better than that of WL-KLD, which is in line with the FROC metrics.}
    \label{fig:result_vis}
\end{figure*}

\subsection{Network Architecture}
\label{sec.method.network}

%Due to the intrinsically different types and natures of 
The radiographic appearances of pelvic fractures can vary considerably in both scale and shape. To represent the pathological patterns at different scales, we employ a ResNet-50 with a \ac{FPN} backbone~\cite{lin2017feature}. The ResNet-50 portion has a bottom-up path that encodes the input \ac{PXR} using four sequential blocks, with each down-sampling the image size by $2$. The FPN adds a top-down path that upsamples and fuses the bottom-up feature maps. The feature maps are then processed by a 1x1 convolution to produce dense probability maps at four different scales for the input PXR image. The output probability maps can be interpreted as the confidence of presence of fractures at the corresponding locations in the \ac{PXR}.

%, resulting in four feature maps $ \{ F_1,F_2,F_3,F_4 \} $ with different spatial resolutions, respectively
%, denoted as $\{P_1,P_2,P_3,P_4\}$

FPNs have been widely adopted in object detection to cope with the variance in object scales using multiple pyramid levels of feature maps with respect to different spatial resolutions. With bounding box annotations, objects are assigned to different pyramid levels according to their size. For fracture detection in PXR with point-based annotations, the size of the fracture is inherently ambiguous (from perspectives of both the complexity of the fractures and the form of annotation). Therefore, we assign every fracture annotation point to all FPN pyramid levels to encourage the network to recognize the fractures at different spatial resolutions. During training, losses are calculated on all pyramid levels. During inference, the probability maps from the four levels are ensembled and merged by averaging. The total training loss is the sum of the Window Loss on the four-level FPN outputs, written as:
\begin{equation}
    L=\sum_{k=1,2,3,4} \E_{i,j} \bigg[ WL\Big( P_k(i,j), \ell_k(i,j), u_k(i,j) \Big) \bigg],
\end{equation}
where $P_k$ is the probability map output at the $k$-th pyramid level, $\ell_k$ and $u_k$ are the confidence bounds resized to the spatial resolution of $P_k$.

\subsection{Implementation Details}

We trained our model on a workstation with a single Intel Xeon E5-2650 v4 CPU @ 2.2 GHz, 128 GB RAM, 4 NVIDIA Quadro RTX 8000 GPUs. All methods are implemented in Python 3.7, and PyTorch v1.4. We use the ImageNet pre-trained weights to initialize the backbone network. The Adam optimizer with a learning rate of $4e-5$, a weight decay of $0.001$ and a batch size of $48$ is used to train the model for $100$ epochs. All images are padded to square and resized to $1024 \times 1024$ for network training and inference. We randomly perform rotation, horizontal flipping, intensity and contrast jittering to augment the training data. The supervision bounds are generated using parameters $r_{\ell}=50$, $r_{u}=200$ and $\tau=2$. The trained model is evaluated on the validation set after every training epoch, and the one with the highest validation \ac{AUROC} is selected as the best model for inference.

\section{Experiments}\label{sec.exp}

%\subsection{Dataset}

{\bf Dataset:} We retrieved \num{4410} PXR images of unique patients that were recorded from 2008 to 2016 in the trauma registry of \textit{anonymous hospital}. Trauma-related findings are identified by a board consisting of a radiologist, a trauma surgeon, and an orthopedic surgeon who had fifteen years, seven years, and three years of experience, respectively. The best available information was provided, including the original images, radiologist reports, clinical diagnoses, surgical reports, and advanced imaging modality findings, if available. A total of \num{2776} images with acute trauma-related radiographic findings (1975 and 801 hip and pelvic fractures, respectively) are identified, resulting in \num{4066} annotated points (range 0-7). All experiments are conducted using five-fold cross-validation with a $70\%/10\%/20\%$ training, validation, and testing split, respectively.

\subsection{Evaluation Metrics}

% We evaluate both the image-level classification and fracture localization performances for all compared methods. 
%The image-level classification performance is critical for trauma \ac{CAD} systems because the presence of any fracture affects the downstream clinical workflow. The fracture localization performance is also of significant clinical interest since it helps physicians to identify fracture sites.

\subsubsection{Classification metric}
We evaluate the image-level fracture classification performance using the \ac{ROC} curve and the widely used \ac{AUROC} classification metric. For methods predicting probability map (e.g., the proposed method), the maximum response of the probability map is taken as the image-level classification score. For object detection methods predicting bounding box, the maximum classification score of all predicted boxes is taken as the image-level classification score. 

\subsubsection{Localization metric}
We evaluate the fracture localization performance of different methods using \ac{FROC} curve. To quantify FROC, we calculate an FROC score as the average recall at five false positive (FP) rates: (0.1, 0.2, 0.3, 0.4, 0.5) FPs per image. We also separately report the recall at FP=0.1 (Recall@0.1), the most clinically relevant setting. Given an operating point of the FROC, the predicted heatmap is binarized using the operating point to produce a mask. Connected component analysis is performed on the mask and each connected component is regarded as one detected fracture finding. A bounding box is then generated on each connected component. An annotation point is considered to be recalled if it is inside any bounding box. To define false positives, a ground truth mask is first generated with disks of radius 50 centered on the annotation points, which indicates areas that are certainly affected by the anomalies. A bounding box is considered as false positive if its intersection with the mask is less than 10\% of its own region. 

\begin{figure}[t]
    \centering
    \noindent\includegraphics[width=0.98\linewidth]{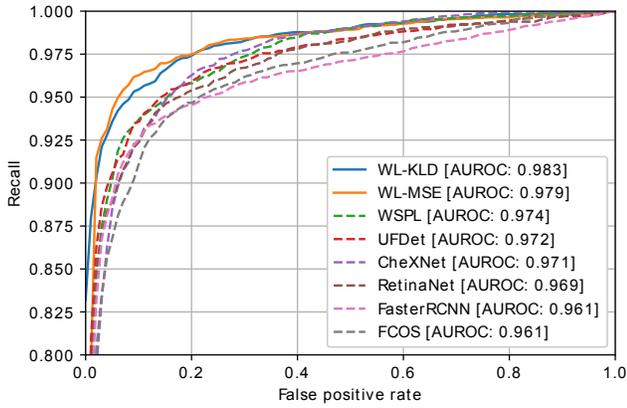} 
    \caption{Comparison of receiver operating characteristic (ROC) curves (Recall scaled from 0.8 to 1.0).}
    \label{fig:auroc}
\end{figure} \vspace{-2mm}
\subsection{Comparison with baseline methods}

\subsubsection{Image classification methods}
We evaluate three state-of-the-art X-ray \ac{CAD} methods based on image classification, \textit{CheXNet} \cite{rajpurkar2017chexnet}, \textit{WSPL}\footnote{Weakly-Supervised Pathology Localization} \cite{wang2017chestx8} and \textit{UFDet}\footnote{Universal Fracture Detection} \cite{wang2019weakly}. CheXNet trains a classification network with a \ac{GAP} layer on the last feature map followed by a fully connected layer. WSPL replaces the \ac{GAP} with a \ac{LSE} pooling. UFDet estimates a dense probability map and use \ac{LSE} pooling to produce the classification score. We evaluate the stage-1 of UFDet for a fair comparison with CheXNet, WSPL and our method, which are all single-stage methods. Anomaly localization map is generated using \ac{CAM} \cite{zhou2016learning} for CheXNet and WSPL, and the produced probability map for UFDet. The localization map is converted to bounding box for FROC evaluation using the same steps described above. ResNet-50 is used as the backbone network for all image classification methods.

\subsubsection{Object detection methods}

We evaluate three state-of-the-art object detection methods including two popular anchor-based detectors, \textit{Faster-RCNN}~\cite{ren2015faster} and \textit{RetinaNet} \cite{lin2017focal}, and the latest anchor-less detector, \textit{FCOS} \cite{tian2019fcos}. All compared methods use ResNet-50 with FPN as the backbone network. Since the actual scale of the fracture is unknown from the point-based annotation, a bounding box with a heuristically chosen size of $200\times200$ is placed on each annotation point. We empirically verified that the size of the bounding box is appropriate for the majority of fractures observed in \acp{PXR}.

\begin{figure}[t]
    \centering
    \noindent\includegraphics[width=0.98\linewidth]{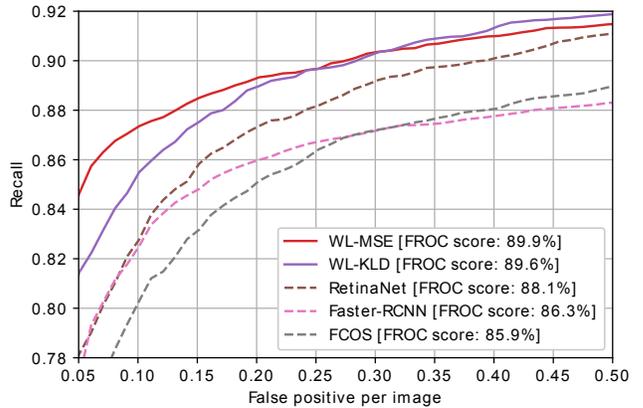} 
    \caption{Comparison of free-response receiver operating characteristic (FROC) curves.}
    \label{fig:froc} \vspace{-2mm}
\end{figure}

\begin{table}[b]
  \centering
  \caption{\textbf{Fracture classification and localization results}. Our method achieves the top performance, outperforming all baseline methods by significant margins.} 
  \begin{tabular}{lp{1.2cm}<{\centering}p{1.5cm}<{\centering}p{1.5cm}<{\centering}}
  \toprule
  Method & AUROC & Recall $@$FP=0.1 & FROC Score \\
  \midrule
  CheXNet     & 0.971  & 28.9\%     & 54.9\%     \\
  WSPL        & 0.974  & 56.0\%     & 68.5\%     \\
  UFDet       & 0.972  & 64.9\%     & 75.4\%     \\ \midrule
  Faster-RCNN & 0.961  & 82.5\%     & 86.3\%     \\
  RetinaNet   & 0.969  & 82.6\%     & 88.1\%     \\
  FCOS        & 0.961  & 80.3\%     & 85.9\%     \\ \midrule
  WL-MSE      & 0.979  & \bf 87.4\% & \bf 89.9\% \\
  WL-KLD      & \bf 0.983 & 85.4\%     & 89.6\%     \\
  \bottomrule
  \end{tabular}
  \label{tab:benchmark} \vspace{-2mm}
\end{table}

% \begin{table}[b]
%   %\renewcommand\arraystretch{1.5}
%   \centering
%   \caption{\textbf{Fracture classification and localization results}. Our method achieves the top performance, outperforming all baseline methods by significant margins.} 
%   \begin{tabular}{lp{0.8cm}<{\centering}p{1.5cm}<{\centering}p{1.5cm}<{\centering}p{0.8cm}<{\centering}}
%   \toprule
%   Method & AUROC & Recall $@$Spec=0.95 & Recall $@$FP=0.1 & FROC Score \\
%   \midrule
%   CheXNet     & 0.971  & 88.3\%   & 28.9\%     & 54.9\%     \\
%   WSPL        & 0.974  & 90.1\%   & 56.0\%     & 68.5\%     \\
%   UFDet       & 0.972  & 90.5\%   & 64.9\%     & 75.4\%     \\ \midrule
%   Faster-RCNN & 0.961  & 88.8\%   & 82.5\%     & 86.3\%     \\
%   RetinaNet   & 0.969  & 88.9\%   & 82.6\%     & 88.1\%     \\
%   FCOS        & 0.961  & 86.7\%   & 80.3\%     & 85.9\%     \\ \midrule
%   WL-MSE      & 0.979  & \bf 93.9\%   & \bf 87.4\% & \bf 89.9\% \\
%   WL-KLD      & \bf 0.983 & 93.4\% & 85.4\%     & 89.6\%     \\
%   \bottomrule
%   \end{tabular}
%   \label{tab:benchmark} \vspace{-2mm}
% \end{table}

\subsubsection{Result analysis}

Table \ref{tab:benchmark} summarizes the results of image classification, object detection and the proposed methods. The proposed methods using both \ac{MSE} and \ac{KLD} as the divergence metric outperform both image classification and object detection methods on both the classification (\ie{} AUROC) and localization (\ie{} Recall@0.1 and FROC score) metrics by large margins. Specifically, WL-KLD achieves an AUROC of 0.983, outperforming the closest competitor, WSPL (0.974), by a health gap of 0.009. On Recall@0.1 and FROC score, WL-KLD leads the strongest baseline method, RetinaNet, by significant margins, \ie{} 2.8\% (85.4\% vs. 82.6\%) and 1.5\% (89.6\% vs. 88.1\%), respectively. WL-MSE achieves even stronger localization performance than WL-KLD, reporting a Recall@0.1 of 87.4\% and a FROC score of 89.9\%. The AUROC of WL-MSE is lower than WL-KLD (97.9\% vs. 98.3\%), but it still outperforms all baseline methods by more than a 0.5\%-1.8\% margin.

\begin{table}[!t]
  \centering
  \caption{\textbf{Ablation study on the Window Loss with MSE}.}
  \begin{tabular}{cccp{1cm}<{\centering}p{1.5cm}<{\centering}p{1.8cm}<{\centering}}
  \toprule
  $\tau$   & $r_{\ell}$  & $r_u$ & AUROC & Recall$@$0.1 & FROC Score \\ \midrule
  2        & 50   & 200 & 0.979  & 87.4\%    & 89.9\%       \\ 
  2        & 100  & 200 & 0.978  & 83.4\%    & 86.4\%       \\
  2        & 150  & 200 & 0.979  & 80.5\%    & 82.0\%       \\\midrule
  0.4      & 50   & 200 & 0.978  & 86.7\%    & 89.3\%       \\
  2        & 50   & 200 & 0.979  & 87.4\%    & 89.9\%       \\
  5        & 50   & 200 & 0.979  & 87.4\%    & 88.2\%       \\
  10       & 50   & 200 & 0.982  & 69.6\%    & 84.3\%       \\ \midrule
  $0$ & 50   & 50  & 0.976  & 81.8\%    & 86.1\%       \\ 
  2 & 50 & 50 & 0.975 & 83.9\% & 87.2\% \\
  \bottomrule
  \end{tabular}
  \label{tab:wl_mse} \vspace{-2mm}
\end{table}

An interesting observation is that although image classification methods do not use localization supervisory signals, they achieve superior classification performances comparing to object detection methods trained using localization supervision signals. In particular, the highest \acp{AUROC} achieved by image classification methods and object detection methods are 0.974 and 0.969, respectively, measuring a gap of 0.005. This suggests that algorithms designed specifically for detecting objects may not be optimal for recognizing the image-level existence of the object. On the other hand, image classification methods in general result in poor localization performance, with \ac{FROC} scores between 54.9\% and 75.4\%, while all three object detection methods report substantially higher FROC scores between 85.9\% and 88.1\%. The high \ac{AUROC} and low \ac{FROC} score indicate that without localization supervision signals, the attention mechanism of image classification methods can activate wrong regions even when the image-level classification is correct. This is also evidenced by the examples shown in Fig. \ref{fig:result_vis}.

While there is a trade-off between classification and localization performances using image classification and object detection methods, our method can simultaneously achieve improved performance in both aspects comparing to methods in both categories. Depending on whether \ac{MSE} or \ac{KLD} is used as the divergence measure in the Window Loss, our method behaves slightly differently in classification and localization, \ie{} \ac{KLD} results in a slightly higher \ac{AUROC} (98.3\% vs. 97.9\%), while \ac{MSE} results in a slight higher \ac{FROC} score (89.9\% vs. 89.6\%).

\subsection{Ablation experiments}

We conduct ablation experiments to analyze the effects of the parameters of the Window Loss. We evaluate our method with varying parameters $r_{\ell}$, $r_u$ and $\tau$ using both \ac{MSE} and \ac{KLD} as the divergence measure. The results are recorded in Table \ref{tab:wl_mse} (using \ac{MSE}) and Table \ref{tab:wl_kldiv} (using \ac{KLD}). 

We first evaluate two degenerate setting $[r_{\ell},r_u=50, \tau=0]$ and $[r_{\ell},r_u=50, \tau=2]$. In the first setting, the supervision bounds degenerate to a binary mask with disks of radius 50, and the Window Losses using MSE and KLD divergence measures degenerate to \ac{MSE} and \ac{BCE}, respectively. In the seconding setting, the supervision signal degenerate to a soft mask, similar to the Gaussian ground truth map widely used in key point localization methods~\cite{wei2016convolutional}. As shown in Table \ref{tab:wl_mse}, both settings result in lower classification and localization performances comparing to our default setting $[r_{\ell}=50, r_u=200, \tau=2]$. It demonstrates the effectiveness of the proposed supervision bounds and the Window Loss. It is also worth noting that the degenerate setting, especially using \ac{KLD}, still performs competitively comparing to image classification and object detection methods, measuring an \ac{AUROC} score of 0.980 and a \ac{FROC} score of 85.4\%. It indicates the merit of modeling the anomaly detection task as a pixel-wise classification task.

To further analyze the effects of the parameters, we evaluate our method using $r_{\ell}=[50, 100, 150]$, $\tau=[0.4, 2, 5, 10]$ and $r_u=200$. We use fixed $r_u=200$ to simplify the analysis since it is empirically confirmed that areas more than 200 pixels away from any annotation points never have anomaly. As $r_{\ell}$ increases from 50 to 150, a bigger neighborhood of the annotation point will have a higher lower bound, increasing the lower disallowed region. The results show that the classification performance remains very stable when $r_{\ell}$ increases, \ie{} the difference among the \ac{AUROC} score is within 0.1. However, the localization performance degrades noticeably as the $r$ increases, with the \ac{FROC} score dropped by 7.9\% (resp., 6.8\%) using \ac{MSE} (resp., \ac{KLD}). We posit that because the larger positive region of a higher $r_{\ell}$ contains more normal areas, training the model to produce positive response in these regions can lower its localization accuracy. On the other hand, the results show that with a smaller $\tau=0.4$, the classification performance remains almost the same as the $\tau=2$ using both MSE and KLD, while the localization performance dropped slightly (0.7\% in the FROC score) using MSE. When a larger $\tau=5$ is used, the localization performance starts to degrade noticeably, \ie{} FROC score from 89.9\% to 88.2\% (resp., from 89.6\% to 86.6\%) using MSE (resp., KLD). As $\tau$ further increases to 10, the localization performance degrades significantly, \ie{} Recall@0.1 from 87.4\% to 69.6\% (resp., from 82.0\% to 69.2\%) using \ac{MSE} (resp., \ac{KLD}). This is because the overly smoothed supervision bounds have high tolerance of false activations in normal areas, preventing the training loss to provide sufficient guidance for localization.

\begin{table}[!t]
  \centering
  \caption{\textbf{Ablation study on the Window Loss with KLD}.}
  \begin{tabular}{cccp{1cm}<{\centering}p{1.5cm}<{\centering}p{1.8cm}<{\centering}}
  \toprule
  $\tau$   & $r_{\ell}$  & $r_u$ & AUROC & Recall$@$0.1 & FROC Score \\ \midrule
  2        & 50   & 200 & 0.983  & 85.4\%    & 89.6\%       \\ 
  2        & 100  & 200 & 0.984  & 81.2\%    & 85.4\%       \\
  2        & 150  & 200 & 0.984  & 82.8\%    & 82.8\%       \\ \midrule
  0.4      & 50   & 200 & 0.982  & 85.5\%    & 89.7\%       \\
  2        & 50   & 200 & 0.983  & 85.4\%    & 89.6\%       \\
  5        & 50   & 200 & 0.984  & 82.0\%    & 86.6\%        \\
  10       & 50   & 200 & 0.986  & 69.2\%    & 80.6\%       \\ \midrule
  $0$ & 50   & 50  & 0.980  & 78.3\%    & 85.4\%       \\ 
  2 & 50   & 50  & 0.981  & 83.0\%    & 87.3\%       \\
  \bottomrule
  \end{tabular}
  \label{tab:wl_kldiv} \vspace{-2mm}
\end{table}

\section{Conclusion}

In this paper, we demonstrate that anomalies or pathologies in X-ray imaging can have complex appearance and anatomical nature, which makes the bounding annotation unreliable. We present a disease detection method with a novel \textit{Window Loss} that utilizes more flexible and efficient point-based annotations (capable of capturing anomaly findings in a nonparametric manner by point sampling). We validate the merit of our point-based annotation and Window Loss on the pelvic X-ray fracture detection task. The proposed method achieves substantially improved performance when comparing to state-of-the-art image classification and object detection methods. We also justify the usage of upper/lower bounds with allowed regions via our ablation study. Future studies should aim to employ the point-base annotation and  Window Loss for detecting a broader range of anomalies in X-ray images from different body parts, as well as extending the method to new imaging modalities (\eg{} CT and MRI).

\bibliography{mybib}

\begin{thebibliography}{34}
\providecommand{\natexlab}[1]{#1}
\providecommand{\url}[1]{\texttt{#1}}
\providecommand{\urlprefix}{URL }
\expandafter\ifx\csname urlstyle\endcsname\relax
  \providecommand{\doi}[1]{doi:\discretionary{}{}{}#1}\else
  \providecommand{\doi}{doi:\discretionary{}{}{}\begingroup
  \urlstyle{rm}\Url}\fi

\bibitem[{Al-Masni et~al.(2018)Al-Masni, Al-Antari, Park, Gi, Kim, Rivera,
  Valarezo, Choi, Han, and Kim}]{al2018simultaneous}
Al-Masni, M.~A.; Al-Antari, M.~A.; Park, J.-M.; Gi, G.; Kim, T.-Y.; Rivera, P.;
  Valarezo, E.; Choi, M.-T.; Han, S.-M.; and Kim, T.-S. 2018.
\newblock Simultaneous detection and classification of breast masses in digital
  mammograms via a deep learning YOLO-based CAD system.
\newblock \emph{Computer methods and programs in biomedicine} 157: 85--94.

\bibitem[{Baltruschat et~al.(2019)Baltruschat, Nickisch, Grass, Knopp, and
  Saalbach}]{baltruschat2019comparison}
Baltruschat, I.~M.; Nickisch, H.; Grass, M.; Knopp, T.; and Saalbach, A. 2019.
\newblock Comparison of deep learning approaches for multi-label chest X-ray
  classification.
\newblock \emph{Scientific reports} 9(1): 1--10.

\bibitem[{Cai et~al.(2020)Cai, Harrison, Zheng, Yan, Huo, Xiao, Yang, and
  Lu}]{cai2020lesion}
Cai, J.; Harrison, A.~P.; Zheng, Y.; Yan, K.; Huo, Y.; Xiao, J.; Yang, L.; and
  Lu, L. 2020.
\newblock Lesion Harvester: Iteratively Mining Unlabeled Lesions and
  Hard-Negative Examples at Scale.
\newblock \emph{IEEE Transactions on Medical Imaging [Accepted]} .

\bibitem[{Chellam(2016)}]{chellam2016missed}
Chellam, W. 2016.
\newblock Missed subtle fractures on the trauma-meeting digital projector.
\newblock \emph{Injury} 47(3): 674--676.

\bibitem[{Chen et~al.(2020)Chen, Wang, Zheng, Li, Chang, Harrison, Xiao, Hager,
  Lu, Liao et~al.}]{chen2020anatomy}
Chen, H.; Wang, Y.; Zheng, K.; Li, W.; Chang, C.-T.; Harrison, A.~P.; Xiao, J.;
  Hager, G.~D.; Lu, L.; Liao, C.-H.; et~al. 2020.
\newblock Anatomy-aware siamese network: Exploiting semantic asymmetry for
  accurate pelvic fracture detection in x-ray images.
\newblock In \emph{European Conference on Computer Vision}, 239--255. Springer.

\bibitem[{Huang et~al.(2017)Huang, Liu, Van Der~Maaten, and
  Weinberger}]{huang2017densely}
Huang, G.; Liu, Z.; Van Der~Maaten, L.; and Weinberger, K.~Q. 2017.
\newblock Densely connected convolutional networks.
\newblock In \emph{Proceedings of the IEEE conference on computer vision and
  pattern recognition}, 4700--4708.

\bibitem[{Huang et~al.(2020)Huang, Liu, Wang, Fang, Wang, Wang, Chen, Chen,
  Meng, and Wang}]{huang2020rectifying}
Huang, Y.-J.; Liu, W.; Wang, X.; Fang, Q.; Wang, R.; Wang, Y.; Chen, H.; Chen,
  H.; Meng, D.; and Wang, L. 2020.
\newblock Rectifying Supporting Regions with Mixed and Active Supervision for
  Rib Fracture Recognition.
\newblock \emph{IEEE Transactions on Medical Imaging} .

\bibitem[{Irvin et~al.(2019)Irvin, Rajpurkar, Ko, Yu, Ciurea-Ilcus, Chute,
  Marklund, Haghgoo, Ball, Shpanskaya et~al.}]{irvin2019chexpert}
Irvin, J.; Rajpurkar, P.; Ko, M.; Yu, Y.; Ciurea-Ilcus, S.; Chute, C.;
  Marklund, H.; Haghgoo, B.; Ball, R.; Shpanskaya, K.; et~al. 2019.
\newblock Chexpert: A large chest radiograph dataset with uncertainty labels
  and expert comparison.
\newblock In \emph{Proceedings of the AAAI Conference on Artificial
  Intelligence}, volume~33, 590--597.

\bibitem[{Jiang et~al.(2020)Jiang, Wang, Liang, Xu, and
  Xiao}]{jiang2020elixirnet}
Jiang, C.; Wang, S.; Liang, X.; Xu, H.; and Xiao, N. 2020.
\newblock ElixirNet: Relation-Aware Network Architecture Adaptation for Medical
  Lesion Detection.
\newblock In \emph{AAAI}, 11093--11100.

\bibitem[{Law and Deng(2018)}]{law2018cornernet}
Law, H.; and Deng, J. 2018.
\newblock Cornernet: Detecting objects as paired keypoints.
\newblock In \emph{Proceedings of the European Conference on Computer Vision
  (ECCV)}, 734--750.

\bibitem[{Li et~al.(2018)Li, Wang, Han, Xue, Wei, Li, and
  Fei-Fei}]{li2018thoracic}
Li, Z.; Wang, C.; Han, M.; Xue, Y.; Wei, W.; Li, L.-J.; and Fei-Fei, L. 2018.
\newblock Thoracic disease identification and localization with limited
  supervision.
\newblock In \emph{Proceedings of the IEEE Conference on Computer Vision and
  Pattern Recognition}, 8290--8299.

\bibitem[{Lin et~al.(2017{\natexlab{a}})Lin, Doll{\'a}r, Girshick, He,
  Hariharan, and Belongie}]{lin2017feature}
Lin, T.-Y.; Doll{\'a}r, P.; Girshick, R.; He, K.; Hariharan, B.; and Belongie,
  S. 2017{\natexlab{a}}.
\newblock Feature pyramid networks for object detection.
\newblock In \emph{Proceedings of the IEEE conference on computer vision and
  pattern recognition}, 2117--2125.

\bibitem[{Lin et~al.(2017{\natexlab{b}})Lin, Goyal, Girshick, He, and
  Doll{\'a}r}]{lin2017focal}
Lin, T.-Y.; Goyal, P.; Girshick, R.; He, K.; and Doll{\'a}r, P.
  2017{\natexlab{b}}.
\newblock Focal loss for dense object detection.
\newblock In \emph{Proceedings of the IEEE international conference on computer
  vision}, 2980--2988.

\bibitem[{Lin et~al.(2014{\natexlab{a}})Lin, Maire, Belongie, Hays, Perona,
  Ramanan, Doll{\'a}r, and Zitnick}]{lin2014microsoft}
Lin, T.-Y.; Maire, M.; Belongie, S.; Hays, J.; Perona, P.; Ramanan, D.;
  Doll{\'a}r, P.; and Zitnick, C.~L. 2014{\natexlab{a}}.
\newblock Microsoft coco: Common objects in context.
\newblock In \emph{European conference on computer vision}, 740--755. Springer.

\bibitem[{Lin et~al.(2014{\natexlab{b}})Lin, Maire, Belongie, Hays, Perona,
  Ramanan, Dollár, and Zitnick}]{Lin2014coco}
Lin, T.-Y.; Maire, M.; Belongie, S.~J.; Hays, J.; Perona, P.; Ramanan, D.;
  Dollár, P.; and Zitnick, C.~L. 2014{\natexlab{b}}.
\newblock Microsoft COCO: Common Objects in Context.
\newblock In \emph{ECCV}, 5:740--755.

\bibitem[{Liu et~al.(2019{\natexlab{a}})Liu, Zhao, Fei, Zhang, Wang, and
  Yu}]{liu2019align}
Liu, J.; Zhao, G.; Fei, Y.; Zhang, M.; Wang, Y.; and Yu, Y. 2019{\natexlab{a}}.
\newblock Align, attend and locate: Chest x-ray diagnosis via contrast induced
  attention network with limited supervision.
\newblock In \emph{Proceedings of the IEEE International Conference on Computer
  Vision}, 10632--10641.

\bibitem[{Liu et~al.(2019{\natexlab{b}})Liu, Zhou, Zhang, Luo, Zhang, Zhang,
  Li, Wang, and Yu}]{liu2019unilateral}
Liu, Y.; Zhou, Z.; Zhang, S.; Luo, L.; Zhang, Q.; Zhang, F.; Li, X.; Wang, Y.;
  and Yu, Y. 2019{\natexlab{b}}.
\newblock From Unilateral to Bilateral Learning: Detecting Mammogram Masses
  with Contrasted Bilateral Network.
\newblock In \emph{International Conference on Medical Image Computing and
  Computer-Assisted Intervention}, 477--485. Springer.

\bibitem[{Papadopoulos et~al.(2017)Papadopoulos, Uijlings, Keller, and
  Ferrari}]{Papadopoulos2017Extreme}
Papadopoulos, D.; Uijlings, J.; Keller, F.; and Ferrari, V. 2017.
\newblock Extreme clicking for efficient object annotation.
\newblock In \emph{ICCV}.

\bibitem[{Pham et~al.(2020)Pham, Le, Ngo, Tran, and
  Nguyen}]{pham2020interpreting}
Pham, H.~H.; Le, T.~T.; Ngo, D.~T.; Tran, D.~Q.; and Nguyen, H.~Q. 2020.
\newblock Interpreting Chest X-rays via CNNs that Exploit Hierarchical Disease
  Dependencies and Uncertainty Labels.
\newblock \emph{arXiv preprint arXiv:2005.12734} .

\bibitem[{Rajpurkar et~al.(2017)Rajpurkar, Irvin, Zhu, Yang, Mehta, Duan, Ding,
  Bagul, Langlotz, Shpanskaya et~al.}]{rajpurkar2017chexnet}
Rajpurkar, P.; Irvin, J.; Zhu, K.; Yang, B.; Mehta, H.; Duan, T.; Ding, D.;
  Bagul, A.; Langlotz, C.; Shpanskaya, K.; et~al. 2017.
\newblock Chexnet: Radiologist-level pneumonia detection on chest x-rays with
  deep learning.
\newblock \emph{arXiv preprint arXiv:1711.05225} .

\bibitem[{Ren et~al.(2015)Ren, He, Girshick, and Sun}]{ren2015faster}
Ren, S.; He, K.; Girshick, R.; and Sun, J. 2015.
\newblock Faster r-cnn: Towards real-time object detection with region proposal
  networks.
\newblock In \emph{Advances in neural information processing systems}, 91--99.

\bibitem[{Sirazitdinov et~al.(2019)Sirazitdinov, Kholiavchenko, Mustafaev,
  Yixuan, Kuleev, and Ibragimov}]{sirazitdinov2019deep}
Sirazitdinov, I.; Kholiavchenko, M.; Mustafaev, T.; Yixuan, Y.; Kuleev, R.; and
  Ibragimov, B. 2019.
\newblock Deep neural network ensemble for pneumonia localization from a
  large-scale chest x-ray database.
\newblock \emph{Computers \& Electrical Engineering} 78: 388--399.

\bibitem[{Tian et~al.(2019)Tian, Shen, Chen, and He}]{tian2019fcos}
Tian, Z.; Shen, C.; Chen, H.; and He, T. 2019.
\newblock Fcos: Fully convolutional one-stage object detection.
\newblock In \emph{Proceedings of the IEEE international conference on computer
  vision}, 9627--9636.

\bibitem[{Wang et~al.(2017)Wang, Peng, Lu, Lu, Bagheri, and
  Summers}]{wang2017chestx8}
Wang, X.; Peng, Y.; Lu, L.; Lu, Z.; Bagheri, M.; and Summers, R.~M. 2017.
\newblock Chestx-ray8: Hospital-scale chest x-ray database and benchmarks on
  weakly-supervised classification and localization of common thorax diseases.
\newblock In \emph{IEEE CVPR}.

\bibitem[{Wang et~al.(2019)Wang, Lu, Cheng, Jin, Harrison, Xiao, Liao, and
  Miao}]{wang2019weakly}
Wang, Y.; Lu, L.; Cheng, C.-T.; Jin, D.; Harrison, A.~P.; Xiao, J.; Liao,
  C.-H.; and Miao, S. 2019.
\newblock Weakly Supervised Universal Fracture Detection in Pelvic X-Rays.
\newblock In \emph{International Conference on Medical Image Computing and
  Computer-Assisted Intervention}, 459--467. Springer.

\bibitem[{Wei et~al.(2016)Wei, Ramakrishna, Kanade, and
  Sheikh}]{wei2016convolutional}
Wei, S.-E.; Ramakrishna, V.; Kanade, T.; and Sheikh, Y. 2016.
\newblock Convolutional pose machines.
\newblock In \emph{Proceedings of the IEEE conference on Computer Vision and
  Pattern Recognition}, 4724--4732.

\bibitem[{Yahalomi, Chernofsky, and Werman(2019)}]{yahalomi2019detection}
Yahalomi, E.; Chernofsky, M.; and Werman, M. 2019.
\newblock Detection of distal radius fractures trained by a small set of X-ray
  images and Faster R-CNN.
\newblock In \emph{Intelligent Computing-Proceedings of the Computing
  Conference}, 971--981. Springer.

\bibitem[{Yan et~al.(2018{\natexlab{a}})Yan, Yao, Li, Xu, and
  Huang}]{yan2018weakly}
Yan, C.; Yao, J.; Li, R.; Xu, Z.; and Huang, J. 2018{\natexlab{a}}.
\newblock Weakly supervised deep learning for thoracic disease classification
  and localization on chest x-rays.
\newblock In \emph{Proceedings of the 2018 ACM International Conference on
  Bioinformatics, Computational Biology, and Health Informatics}, 103--110.

\bibitem[{Yan et~al.(2020)Yan, Cai, Harrison, Jin, Xiao, and
  Lu}]{yan2020universal}
Yan, K.; Cai, J.; Harrison, A.~P.; Jin, D.; Xiao, J.; and Lu, L. 2020.
\newblock Universal Lesion Detection by Learning from Multiple Heterogeneously
  Labeled Datasets.
\newblock \emph{arXiv preprint arXiv:2005.13753} .

\bibitem[{Yan et~al.(2019)Yan, Tang, Peng, Sandfort, Bagheri, Lu, and
  Summers}]{yan2019mulan}
Yan, K.; Tang, Y.; Peng, Y.; Sandfort, V.; Bagheri, M.; Lu, Z.; and Summers,
  R.~M. 2019.
\newblock Mulan: Multitask universal lesion analysis network for joint lesion
  detection, tagging, and segmentation.
\newblock In \emph{International Conference on Medical Image Computing and
  Computer-Assisted Intervention}, 194--202. Springer.

\bibitem[{Yan et~al.(2018{\natexlab{b}})Yan, Wang, Lu, and
  Summers}]{yan_2018_deeplesion}
Yan, K.; Wang, X.; Lu, L.; and Summers, R.~M. 2018{\natexlab{b}}.
\newblock DeepLesion: automated mining of large-scale lesion annotations and
  universal lesion detection with deep learning.
\newblock \emph{J. Med Imaging} 5(3).

\bibitem[{Yao et~al.(2018)Yao, Prosky, Poblenz, Covington, and
  Lyman}]{yao2018weakly}
Yao, L.; Prosky, J.; Poblenz, E.; Covington, B.; and Lyman, K. 2018.
\newblock Weakly supervised medical diagnosis and localization from multiple
  resolutions.
\newblock \emph{arXiv preprint arXiv:1803.07703} .

\bibitem[{Zhao et~al.(2019)Zhao, Wu, Chen, and Li}]{zhao2019automatic}
Zhao, S.; Wu, X.; Chen, B.; and Li, S. 2019.
\newblock Automatic Vertebrae Recognition from Arbitrary Spine MRI Images by a
  Hierarchical Self-calibration Detection Framework.
\newblock In \emph{International Conference on Medical Image Computing and
  Computer-Assisted Intervention}, 316--325. Springer.

\bibitem[{Zhou et~al.(2016)Zhou, Khosla, Lapedriza, Oliva, and
  Torralba}]{zhou2016learning}
Zhou, B.; Khosla, A.; Lapedriza, A.; Oliva, A.; and Torralba, A. 2016.
\newblock Learning deep features for discriminative localization.
\newblock In \emph{Proceedings of the IEEE conference on computer vision and
  pattern recognition}, 2921--2929.

\end{thebibliography}

\end{document}